\documentclass{acmsiggraph}
\usepackage[ruled]{algorithm2e}
\usepackage{lipsum}
\usepackage{enumitem}
\usepackage{booktabs}
\usepackage{authblk}

\title{Mask-off: Synthesizing Face Images in the Presence of Head-mounted Displays}

\author[1]{Yajie Zhao\thanks{yajie.zhao@uky.edu}}
\author[1]{Qingguo Xu\thanks{qingguo.xu@uky.edu}}
\author[2]{Xinyu Huang\thanks{xinyu.huang@nccu.edu}}
\author[1]{Ruigang Yang\thanks{ryang@cs.uky.edu}}
\affil[1]{University of Kentucky}
\affil[2]{North Carolina Central University}
\pdfauthor{Yajie Zhao}

\TOGonlineid{45678}

\setcopyright{none}

\begin{document}


 \teaser{
   \includegraphics[height=1.4in]{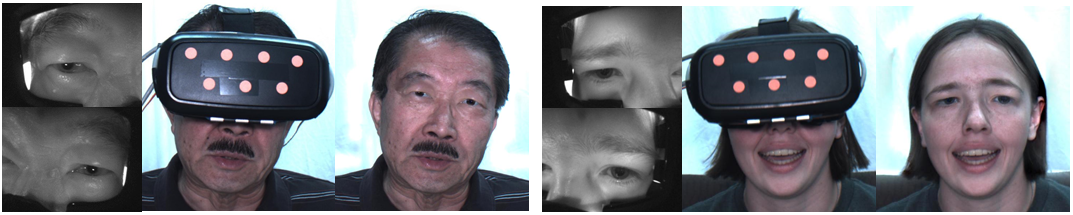}
   \caption{Our system automatically reconstruct photo-realistic face videos for users wearing HMD. (\textit{Left}) Input NIR eye images. (\textit{Middle}) Input face image with upper face blocked by HMD device. (\textit{Right}) The output of our system.  } \label{fig:teaser}
 }

\maketitle

\begin{abstract}

A head-mounted display (HMD) could be an important component of augmented reality system. However, as the upper face region is seriously occluded by the device, the user experience could be affected in applications such as telecommunication and multi-player video games. In this paper, we first present a novel experimental setup that consists of two near-infrared (NIR) cameras to point to the eye regions and one visible-light RGB camera to capture the visible face region. The main purpose of this paper is to synthesize realistic face images without occlusions based on the images captured by these cameras. To this end, we propose a novel synthesis framework that contains four modules: 3D head reconstruction, face alignment and tracking, face synthesis, and eye synthesis. In face synthesis, we propose a novel algorithm that can robustly align and track a personalized 3D head model given a face that is severely occluded by the HMD. In eye synthesis, in order to generate accurate eye movements and dynamic wrinkle variations around eye regions, we propose another novel algorithm to colorize the NIR eye images and further remove the ``red eye" effects caused by the colorization. Results show that both hardware setup and system framework are robust to synthesize realistic face images in video sequences.
\end{abstract}

%
%


%
%




\printcopyright

\section{Introduction}
With the recent surge of interests in virtual reality (VR) and augmented reality (AR) techniques, it is increasingly common to see people wearing head-mounted displays. Often taunted as a new means for social interactions, the form factor of these HMDs, however, severely limit one common form of interactions, that is face-to-face communications, either in the same physical space, or connected via imaging techniques (e.g., video teleconferencing).  In the foreseeable future, HMDs that offer an immersive or seamless experience will severely occlude a large portion of the face.  As a result, it is difficult or even impossible for other people to identify the user, facial expression, and eye gazes.

We are certainly not the first to identify this problem. There are several recent research papers that aim to address this problem. They can roughly be divided into two categories. The first is to find ways to track the expression, using cameras or other sensing devices embedded inside the helmet such as in ~\cite{olszewski2016high,li2015facial,thies16FaceVR}. The tracked expression is then used to drive an avatar. While very impressive results, in both the tracking accuracy and the realism of the final synthesized face images, have been demonstrated, approaches in this category are limited to providing a talking head experience, body movement and gestures, which are also important for communications are missing. The second category aims to inpaint the occluded part, making it possible to present the full picture as if the subject is not wearing the HMD at all. This approach is quite difficult since the occluded part is significant and we are very sensitive to artifacts on the face. We have found only one paper following this direction. In ~\cite{burgos2015real}, Burgos \textit{et al.} first train a regression model of a subject's expressions, then based on the unclouded part (e.g., face) to synthesize a complete face image with expressions. Limited results have been presented.

In this paper, we present a novel framework in the second category to digitally remove the HMD. We use a main stationary camera to capture the subject, as in a typical video conference setup. In addition, we add two small near-IR cameras inside the HMD to track the eye's motion. The goal is to synthesize face part occluded in the main camera image, including pasting the eye images to the correct position and restoring hairline compressed by the straps. To do that, we first build a personalized 3D face model of the subject by using structure-from-motion and parametric face modeling techniques. Using this parametric face model, we precisely track the subject's head pose and expressions at run-time. The tracked expression information, eye images, as well as images of the subject without wearing the HMD,  are used together to fill in the occluded part.

Since we are literately putting different pieces of face parts together, accuracy is of paramount importance. The most significant innovation in this paper is our novel system calibration, tracking, and image warping techniques. In addition we have developed a novel method to colorize eye regions synthesized from NIR cameras and refine them by removing ``red eye" effects. Our method is superior than standard red-eye removal method since the eye image is captured under near-IR illumination in which the eye actually appears differently than from regular illumination. As shown in Figure~\ref{fig:teaser}, our system has been able to produce photo-realistic results.

The rest of the paper is organized as follows. In Section~\ref{sec:related}, we discuss related works. Section~\ref{sec:System1} presents our hardware setup and the method we used to calibrate our system. Our framework and algorithms are described in Section~\ref{sec:overview},~\ref{sec:headmodel_reconstruction},~\ref{sec:alignment},~\ref{sec:face_synthesis},~\ref{sec:eye_synthesis}. Experimental results and conclusions are given in Section~\ref{sec:exp} and~\ref{sec:conc} respectively. 
\section{Related Work}\label{sec:related}


Analysis and synthesis of face expressions have been studied in the past few decades. There are various algorithms have been proposed. For instance, active appearance models (AAM) and 3D morphable models have been successfully used in many applications to model shape and texture variations of faces~\cite{xiao2004real,blanz1999morphable}. Depth sensors are also widely used to reconstruct 3D face models in recent years. Cai \textit{et al.} presented a deformable model fitting algorithm to track 3D face model using a commodity depth camera~\cite{cai20103d}. In~\cite{thies2015real}, Thies \textit{et al.} proposed an reenactment algorithm to transfer facial expressions from one person to another in real-time assuming no occlusions present. A RGB-D sensor is used to estimate and track face model, head pose, and illumination.  Thies \textit{et al.} further extend their reenactment to only use RGB sensor in ~\cite{thies2016face}. It is easy to find out that most of these algorithms are not designed for HMDs. Therefore, they often have either different inputs or outputs comparing with our system and algorithms. For example, Some of them may not be robust under serious occlusions. Some of them output animations of 3D face models, in which eye gazes are not important.

A few research has been done recently to drive 3D avatars for users with HMD. In~\cite{romera2014facial}, Romera-Paredes \textit{et al.} adopted an experimental setup that has two visible light cameras which pointing towards eye regions from oblique
 angles to capture the eye movement. They built a regression framework from the the captured partial face images as input to the blending weights of personalized blendshapes. Multiple machine learning algorithms, such as ridge regression and convolutional neural networks are applied in their framework. In~\cite{li2015facial}, Li \textit{et al.} develops a novel HMD that uses electronic materials (strain sensor) to measure the surface strain signals and RGB-D camera to track visible face regions. A linear mapping is trained between the blendshape coefficients of the whole face and the vector that concatenates strain signals and the blendshape coefficients of the visible face part. This mapping is then used to animate virtual avatars. In ~\cite{olszewski2016high}, they propose an approach for 3D avatar control based on RGB data in real-time. The mouth and eyebrow motion is captured separately by an external camera which attached to HMD and two IR cameras mounted inside the HMD. This data is further used to train a regression model that maps the inputs to a set of coefficients of a parametric 3D avatar. All the above approaches are aimed at 3D parametric avatar driven for HMD users.

Recently, In ~\cite{thies16FaceVR}, Thies \textit{et al.} propose a novel approach for real-time gaze-aware facial reenactment for user's with HMD. They use a RGB-D camera to capture and track the mouth motion and two internal IR cameras to track the eye gazes. They then reconstruct 2D face images by using the tracked expression and pre-recorded videos of the user with eye-gazed corrected. Although they can reconstruct the full face in photo-realistic videos by self-reenactment,their result will lose information in original inputs, like head pose, gestures and other information. In ~\cite{burgos2015real}, Burgos \textit{et al.} build a system to reconstruct face with HMD in 2D videos by using a personalized textured 3D model. They use a RGB camera to capture  face videos and track the expressions, then the textured model is projected and blended with the remaining mouth part. Their approach do not handle eye gaze changes. In our  approach, we will reconstruct HMD faces in photo-realistic videos with ground truth eye movement and the visible information in original input will be faithfully reserved.

In our proposed algorithm, we also integrate various techniques such as landmark detection~\cite{cao2014face}, colorization~\cite{reinhard2001color,levin2004colorization,shih2014style}, and feature extraction~\cite{ke2004pca}. Details are provided in corresponding sections.
\section{Hardware Setup and Calibration}\label{sec:System1}
\subsection{Hardware Setup}\label{sec:hardware}
\begin{figure}[ht]
  \centering
  \includegraphics[width=3.0in]{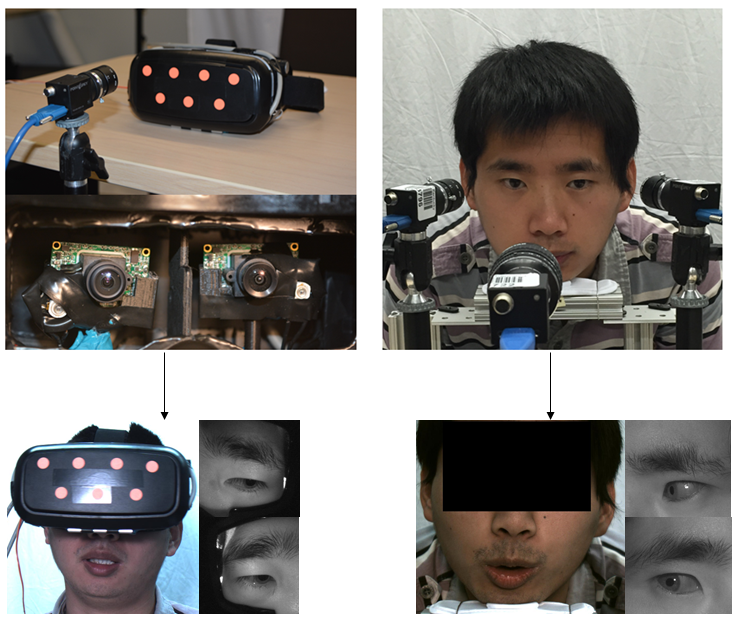}
  \caption{Two experimental systems we have built. (Left Column)our mobile setup. Two small cameras are inside the VR-display case. (Right Column) our simulation setup with three cameras, two for eyes and one for the entire face. The occlusion on face is synthesized by applying a mask.  }
  \label{fig:systemSetup}
\end{figure}

We have built two prototypes for experiments and validation. Our first system is a fixed setup, as shown in Figure~\ref{fig:systemSetup} right. It consists of three cameras. The middle one is a color camera with a resolution of $1280\times960$, it is used to capture the entire face. The other two cameras are near-infrared (NIR) VGA ($640\times480$) cameras capturing the two eye regions using narrow field-of-view lenses. IR LEDs are used to provide sufficient illuminations for the NIR cameras. All cameras are synchronized. The color image can capture the full face of the user without any occlusion. We then simulate the occluded face image by generating masks to block the upper face.  In this setup, the three cameras are used to simulate the case in which all three cameras are rigidly attached to the HMD display,so that head pose doesn't need to be tracked and always be frontal. This setup allows us to capture ground truth images for evaluation purpose. To use this setup, the user is expected to put her/his face on the chin reset to maintain the relative transformation between her/his face and all the cameras.

The second system is a mobile one (Figure~\ref{fig:systemSetup} Left). We use a VR-headset case, one of these types that allow a user to insert a mobile phone to create a low-cost head-mounted display (HMD). We insert two small VGA cameras inside the shell to observe the eye region. It should be noted that since our camera/LED set is not small enough, we do not have the phone inserted during all of our experiments. This limitation could certainly be solved by better (and more costly) engineering. In this mobile setup, a user should wear our modified headset as usual, a fixed RGB camera is used to observed the user. This is similar to a regular video conference setup except that the user's face is severely occluded physically and can move freely in any poses.

%
%
%

\subsection{Calibration}\label{sec:sysCalib}
\begin{figure}[ht]
  \centering
  \includegraphics[width=3.2in]{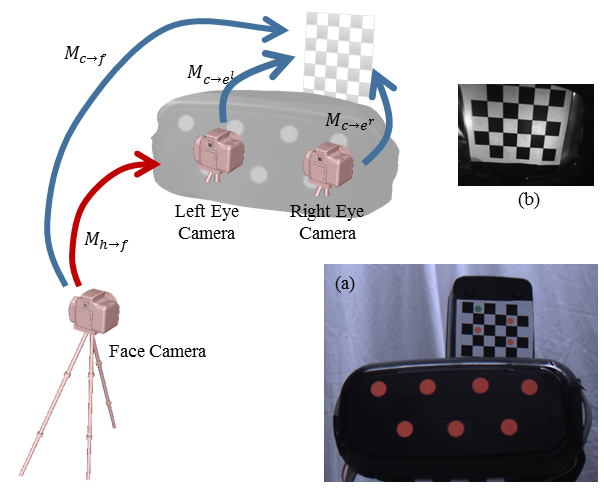}
  \caption{\textit{An illustration of our calibration procedure for the mobile setup}. One checkerboard is placed behind the HMD. The transformations between different cameras/coordinate systems are labeled. Inset (a) shows an image captured by the face camera and inset (b) shows an image captured by one of the eye cameras.  }
  \label{fig:calibrationSystem}
\end{figure}

We first have to geometrically calibrate all cameras. While the fixed setup is easy to deal with, the mobile one is more difficult since the HMD with two internal IR cameras can move freely. We describe our procedure for the mobile system calibration and tracking. We first intrinsically calibrate all the cameras using standard techniques. We then print out a small checkerboard pattern and attached it to the VR-display case so that one half of the patterns are visible to the face camera and the other half is visible to the NIR eye camera. Since the size of the grid is known, we can estimate the pose of these cameras using a Perspective-n-point algorithm (PnP) (e.g.,~\cite{lepetit2009epnp}). Let's denote the points on the checkerboard pattern as $\mathbf{X}_c$ and the relative poses of the face camera and the eye cameras as $M_{c\rightarrow f}$, $M_{c\rightarrow e^l}$, and $M_{c\rightarrow e^r}$ respectively. Furthermore, we put a number of color dots on the front side of the VR-display case. These dots are used for tracking.  They are co-planar and their relative positions are measured. These points, denoted as $\mathbf{X}_h$, define their own coordinate space. Using PnP, the face camera's pose $M_{h\rightarrow f}$ in the HMD ($\mathbf{X}_h$) space can be estimated. Using the face camera as a bridge, we can now calculate the eye-camera's pose in the space of $\mathbf{X}_h$. For the left eye, it is $M_{c\rightarrow e^l}M^{-1}_{c\rightarrow f}M_{h\rightarrow f}$. Now we can remove the checkerboard pattern (since it will occlude the eye cameras). At run time, the face camera will track the HMD's pose using these color dots and therefore the pose of the eye cameras. The involved coordinate transforms and a photo of our calibration patterns are shown in Figure~\ref{fig:calibrationSystem}.

\section{System Overview}\label{sec:overview}
\begin{figure*}[ht]
  \centering
  \includegraphics[width=0.95\textwidth]{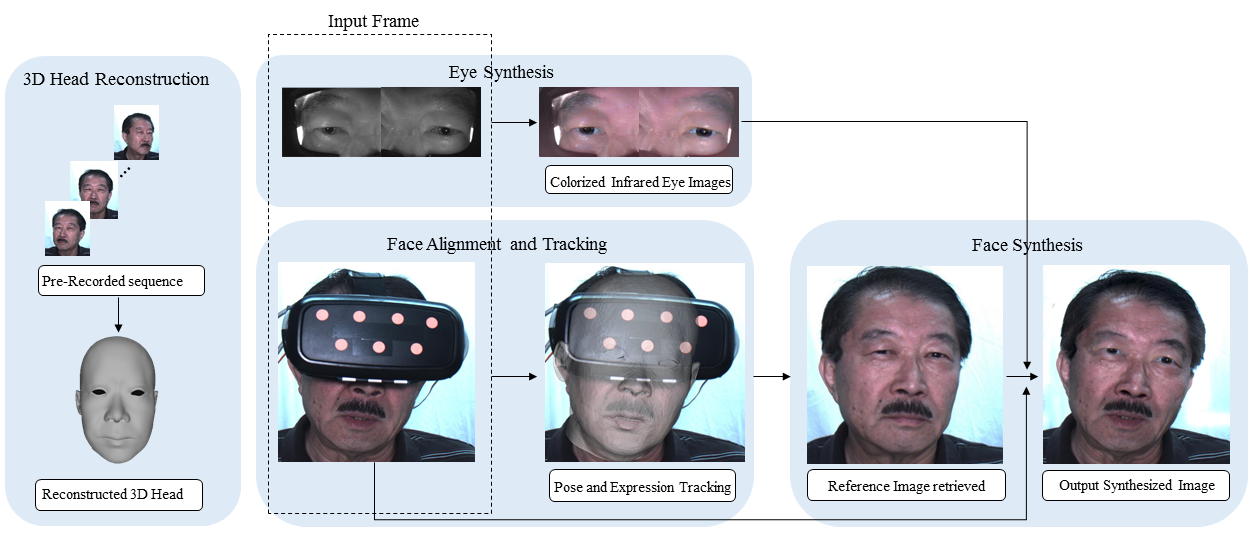}
  \caption{Conceptual overview of our system. From the first stage to the fourth stage, our goal is to synthesize a photo-realistic face image without occlusion.}
  \label{fig:pipeline}
\end{figure*}

Our system consists of four modules as shown in Figure~\ref{fig:pipeline}. We reconstruct a personalized 3D head model from a video sequence captured off-line in the first module (Section~\ref{sec:headmodel_reconstruction}). The 2D facial landmarks and 3D sparse point cloud are integrated together in our optimization algorithm to obtain an accurate head model. In the second module (Section~\ref{sec:alignment}), we propose a novel algorithm to align 3D head model to the face image that has been severely occluded by the HMD. Instead of fitting the head model to the small lower face portion for each image frame, our algorithm first estimates the transformation between the HMD and the head model once a user put on the HMD. The transformation is combined with the estimated HMD pose for each image frame to align the head model robustly. The facial expression weights are then computed to obtain a personalized head model with expression changes. In order to generate realistic face images without occlusions, in the third module (Section~\ref{sec:face_synthesis}), we apply a boundary constrained warping algorithm based on the reference image retrieved from the pre-captured data set. In the fourth module (Section~\ref{sec:eye_synthesis}), we propose another novel algorithm to process the warped near-infrared eye images. The eye images are first colorized based on the color information from the image template. The obvious artifacts (e.g., ``red eye") in the eye regions also are removed in this module.

\section{3D Head Reconstruction}\label{sec:headmodel_reconstruction}

In the off-line data acquisition stage, we record a video sequence of a user with neutral expression under various head poses. The image frames are used to reconstruct a personalized 3D head model for the user. We first apply the structure from motion (SfM) to estimate a sparse point cloud and projection matrices~\cite{hartley2003multiple}. A bi-linear face model described in ~\cite{cao2014facewarehouse} is then used to reconstruct a dense 3D model $M$ with $11K$ vertices from the sparse point cloud,

\begin{equation}\label{eq:multi_linear}
M=B \times_2 C_{id} \times_3 C_{exp},
\end{equation}

where $B\in \mathcal{R}^{11K \times 50 \times 25}$ is the reduced core tensor, $C_{id} \in \mathcal{R}^{50}$ and $C_{exp} \in \mathcal{R}^{25}$ stand for the column vectors of identity weights and expression weights respectively. As we assume the neutral expression during the reconstruction, the expression weights $C_{exp}$ are fixed and only identity weights $C_{id}$ are estimated.

Denote that the reconstructed sparse 3D point cloud as $M^s$, our fitting energy function is defined as,
\begin{equation}
\label{eq:fitting_standalone}
E=\sum_{k=1}^N \Arrowvert sR M_{k}+t-M^s_{k} \Arrowvert^2,
\end{equation}
where the 3D rigid transformation between the sparse point cloud and the bi-linear face model consists of a scale factor $s$, a 3D rotation matrix $R$ and a translation vector $t$. $M_k$ and $M^s_k$ are the $k$th pair of 3D vertices in the dense 3D head model and sparse point cloud. In each iteration, $N$ vertices are selected from the spare point cloud and the corresponding nearest vertices in the dense head model are updated. The initial transformation is computed by using seven 3D facial landmarks reconstructed in 3D point cloud.

We further improve the reconstruction accuracy by using 2D facial landmarks in images that are detected based on the real-time algorithm proposed in~\cite{kazemi2014one}. The cost function is defined as,
\begin{equation}
\label{eq:fitting_2d_images}
E=\sum_{i=1}^{N}\sum_{j=1}^{K}\Arrowvert P_i M_j-l_{ij} \Arrowvert^2+\lambda \sum_{i=1}^{50}((C_{0}^i-C_{id}^i)/\theta)^2
\end{equation}
where $N$ image frames and $K$ facial landmarks in each frame are used. $M_j$ is the $j$th 3D facial landmark in the dense head model, $l_{ij}$ is the $j$th facial landmark in the $i$th image frame, and $P_i$ is the projection matrix for the $i$th image frame. The second term in the equation~\ref{eq:fitting_2d_images} is a regularization term that makes the estimated head model $M$ close to the head model estimated from equation~\ref{eq:fitting_standalone}, which is denoted as $C_{0}$. This term also prevents the the geometry from degeneration and local minima.  




\section{Face Alignment and Tracking}\label{sec:alignment}
As the face is severely occluded by the HMD, the alignment could be highly inaccurate if we align the 3D head model with the face image directly according to remaining visible facial features. In this section, we present a novel approach based on our hardware configuration and calibration.

\subsection{Facial Landmark Detection}
As we need to use facial landmarks in our alignment, three landmark detectors are trained on occluded face image and eye images separately. We use 5 landmarks for eyebrow and 6 landmarks for eye boundary in each eye image, 20 landmarks for mouth, 5 landmarks for nose base and 11 landmarks for lower face boundary in the occluded face image. The cascaded learning framework described in~\cite{cao2014face} is applied. In this learning framework, simple pixel-difference features are extracted and two-level boosted regression is applied. In the internal level of the regression, a set of primitive regressors (\textit{e.g.}, ferns) are trained. Few thousands of training eye images are obtained by cropping labeled face images of the LFW data set~\cite{LFWTech}. As the eye is often located in the middle of the captured image and almost has no in-plane rotations in our hardware setup, we require only few initial eye shapes to achieve accurate predications.

\begin{figure}[ht]
  \centering
  \includegraphics[width=0.3\textwidth]{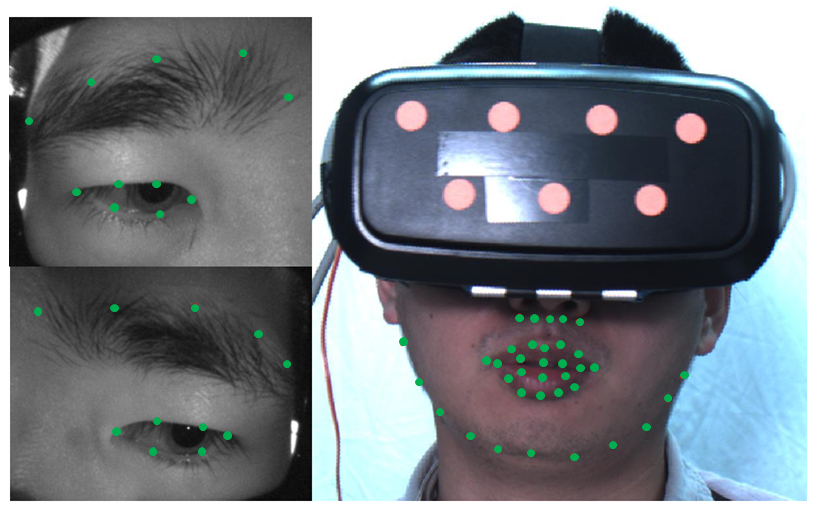}
  \caption{Illustration of the landmarks set we adopted in our system. }
  \label{fig:landmarks}
\end{figure}

\subsection{Initial Alignment}
After the offline calibration described in section 3.2, it is robust to track the HMD's pose (e.g., $P_{h\rightarrow f}$) in real time. The transformations (e.g., $M_{e_l\rightarrow h}$ and $M_{e_r\rightarrow h}$) between eye cameras and the HMD are also fixed after the calibration. However, the transformation between the head and the HMD is different for different users. Even for the same user, the transformation could also be different every time when they wear the HMD. Therefore, it is necessary to estimate the transformation (represented as rotation $R_*$ and translation $T_*$) between the 3D head model and the HMD after a user puts on the device.

In our system, we conduct an initial alignment right after a user puts on the HMD. The user is instructed to change head pose with a neutral expression. The alignment is formulated as a non-linear minimization problem. The cost function $E_{init}$ consists of two terms as shown in Equation~\ref{eq:cost_initial}.
\begin{equation}\label{eq:cost_initial}
E_{init}= E_f+\lambda E_e
\end{equation}
where $\lambda$ is the weight to control $E_e$. The first term $E_f$ is the projection error between visible facial features and corresponding 3D points of the head model. This term is defined in the following Equation.
\begin{equation}\label{eq:initial_term1}
E_f = \sum_{i}d(\mathbf{x}_i,P_{h\rightarrow f}R_*T_*\mathbf{X}_i)^2
\end{equation}
where $\mathbf{x}_i$ and $\mathbf{X}_i$ are the 2D visible facial landmarks in the $i$th image frame of the face camera and corresponding 3D points of the head model, $d(\cdot)$ represents the Euclidean distance between two 2D image points, and $P_{h\rightarrow f}$ is the projection matrix from the HMD device to the face camera. The second term $E_e$ is defined as
\begin{equation}\label{eq:initial_term2}
\begin{array}{cc}
  E_e= & \sum_{i}d(\mathbf{x}_i,P_{h\rightarrow f}M_{e^l\rightarrow h}R_*T_*\mathbf{X}_i)^2 + \\
   & \sum_{i}d(\mathbf{x}_i,P_{h\rightarrow f}M_{e^r\rightarrow h}R_*T_*\mathbf{X}_i)^2
\end{array}
\end{equation}
where $\mathbf{x}_i$ and $\mathbf{X}_i$ are the 2D visible facial landmarks in the $i$th image frame of the NIR eye camera and corresponding 3D points of the head model, and $M_{e^l\rightarrow h}$ and $M_{e^r\rightarrow h}$ are the transformation matrices from eye cameras to the HMD device.

The initial guess of $R_*$ is set to the identity matrix as the rotation between the HMD device and the head model is often very small, and the initial guess of the translation vector in $T_*$ is set to $[0,0,dz]$, where $dz$ is the rough distance between the eye region and the corresponding near infrared camera. The 3D point $\mathbf{X}_i$ on the head model is computed by finding the nearest neighbor of the intersection point between the head model and the ray back projected from $\mathbf{x}_i$. The Levenberg-Marquardt iteration method is applied to solve the cost function.

\subsection{Real-time Alignment}
With the initial alignment, we can easily track the head pose in real time by estimating the projection matrix $P_{h\rightarrow f}$ for each image frame. In this step, we further estimate the expression weights $C_{exp}$ in the bi-linear face model described in Equation~\ref{eq:multi_linear}. Note that the identity weights $C_{id}$ are fixed in this step. The expression weights are estimated based on the energy function,

\begin{equation}\label{eq:cost_exp}
E_{exp}= E_f+\lambda_1 E_e+\lambda_2 E_t+\lambda_3 E_s,
\end{equation}
where $E_f$ and $E_e$ are defined in Equation~\ref{eq:initial_term1} and~\ref{eq:initial_term2} with $\mathbf{X}_i$ replaced by the bi-linear model and the transformation matrices ($R_*$ and $T_*$) fixed. $E_t$ is the constraint imposed by the expression weights from the previous image frame. $E_t$ is defined by,
\begin{equation}\label{eq:cost_exp_et}
E_t(C_{exp})=\parallel C_{exp}^{t-1}-C_{exp}^{t} \parallel^2,
\end{equation}
where $C_{exp}^{t}$ and $C_{exp}^{t-1}$ are the expression weights for current and previous image frame respectively. $E_s$ is the regularization term that restricts the expression weights to to be in the statistical center which avoid of degeneration. $E_s$ is defined as,
\begin{equation}\label{eq:cost_exp_es}
E_s= \sum_{i=1}^{N=25}(C_{exp,i}/\mathbf{\theta}_i)^2
\end{equation}
where $\mathbf{\theta}$ is the mean vector. $E_s$ can also be defined as a Tikhonov regularization energy term $C_{exp}^T D C_{exp}$ with $D=diag(1/\mathbf{\theta}^2)$. The energy function $E_{exp}$ is minimized by the least squares method in real time. The weights we used to balance the terms in our setup is $\lambda_1=2$, $\lambda_2=2$, and $\lambda_3=0.7$.

Our alignment and tracking algorithm is summarized in Algorithm~\ref{alg:headpose}.
\begin{algorithm}
\DontPrintSemicolon
\KwData{Image frames captured by three cameras and a personalized 3D head model.}
\KwResult{Alignment between the 3D model and image frames, $P_{h\rightarrow f}$, $R_*$, $T_*$, $C_{exp}$}

\begin{enumerate}[leftmargin =\parindent, labelwidth =0pt, listparindent = \parindent]
  \item[] \textbf{Initial Alignment}
\end{enumerate}

\begin{enumerate}
  \item Estimate $P_{h\rightarrow f}$ for each time frame based on the color dots.
  \item Facial landmark detection on both occluded face image and eye images.
  \item Initialize $R_*$ to the identity matrix and the translation vector to $[0,0,dz]$.
  \item Back project facial landmarks $\mathbf{x}_i$. Corresponding $\mathbf{X}_i$ are computed by finding the nearest neighbor of intersection with the 3D head model.
  \item Estimate $R_*$ and $T_*$ by minimizing cost function in Equation~\ref{eq:cost_initial}.
  \item Apply transformation to the 3D head model with new $R_*$ and $T_*$ and go to step 4, until converge.
\end{enumerate}

\begin{enumerate}[leftmargin =\parindent, labelwidth =0pt, listparindent = \parindent]
  \item[] \textbf{Real-time Alignment}
\end{enumerate}

\begin{enumerate}
  \item Estimate $P_{h\rightarrow f}$ for each time frame based on the color dots.
  \item Facial landmark detection on both occluded face image and eye images.
  \item Estimate $C_{exp}$ for each time frame by minimizing cost function in Equation~\ref{eq:cost_exp}.
\end{enumerate}

\caption{Head Pose and Expression Tracking\label{alg:headpose}}
\end{algorithm}

\section{Face Synthesis}\label{sec:face_synthesis}

In our face synthesis, we first extract a template image from the data set we have captured off-line, which contains similar head pose as the query image. Then we apply a two-step warping to warp both the template image and the NIR eye images. This method mainly fills the blocked face region with visible region unchanged.

\subsection{Retrieval of Reference Image}
The similarity between $i$th image in the data set and the query image is measured based on three distances as shown in Equation~\ref{eq:retrieval}. The first term is the distance between head poses of the query image ($H_q$) and the reference image candidate ($H_r^i$). The head pose is measured by pitch, yaw, and roll angles based on the transformation $P_{h\rightarrow f}R_*T_*$ that is described in Section 6. The second term is the distance between 2D facial landmarks of the selected reference image in previous time frame ($L_{r-1}$) and current reference image candidate ($L_r^i$). This term removes large 2D translation between two consecutive image frames even they have similar poses. The third term is defined so that current image candidate ($S_r^i$) and previous reference image ($S_{r-1}^i$) have close time stamps. This term could further make the selected reference images continuous.
\begin{equation}\label{eq:retrieval}
D=\parallel H_q-H_r^i \parallel^2 + w_1 \parallel L_r^i-L_{r-1} \parallel^2 + w_2 \parallel S_r^i-S_{r-1} \parallel^2,
\end{equation}
where $w_1$ and $w_2$ are the weights for the second and third term respectively.

\subsection{Face and Eye Image Warping}
\begin{figure}[ht]
  \centering
  \includegraphics[width=0.45\textwidth]{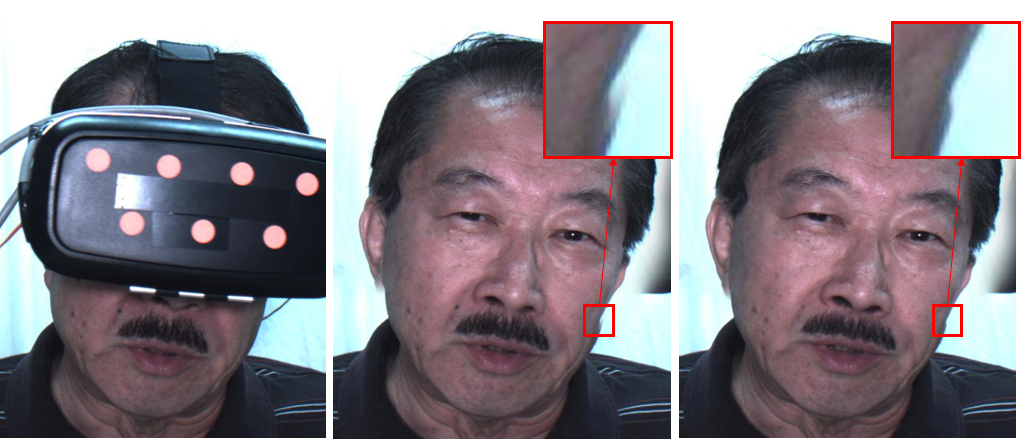}
  \caption{Illustration of the effect of silhouette constraints. (\textit{Left}) is the input target frame needs to be reconstructed. (\textit{Middle}) is the blending result by warping the template without term $E_h$. (\textit{Right} )is the blending result by warping the template with term $E_h$. The result shows that this term forces the warped template image and target image to align on the boundary to eliminate the artifacts. }
  \label{fig:Boundary}
\end{figure}

As the 3D head model have been estimated and aligned with both the reference image and the query image, we project 3D head models to generate dense 2D face meshes. The face mesh for the reference image is then warped to the face mesh for the query image. In order to warp the reference image naturally without obvious distortions, we divide the image to $n\times m$ uniform grid mesh. The energy function is defined as,
\begin{equation}\label{eq:warping}
E= E_d+\alpha E_s +\beta E_b+ \gamma E_h
\end{equation}
where $E_d$ is the data term that assumes bilinear interpolation coefficients remain unchanged after warping, $E_s$ is similarity transformation term based on two set of mesh points, $E_b$ is the term to reduce the transformation outside the face region. Details of these three terms can be found in~\cite{zhao2014video}. As we also want to align the silhouette of the warped template image with the query image to avoid artifacts on the face boundary when blending the warped image and the query image. Therefore, we introduce another term $E_h$ to constrain the silhouette. Denote $\hat{P_{s}}$ and $P_s$ as a pair of 2D correspondence points on silhouette of template image and query image. The template image is divided into $n \times m$ uniform grid mesh $\hat{V}$, the warping problem is to find warped version $V$ of this grid mesh. Then $E_h$ can be formulated as bellow:

\begin{equation}\label{eq:warping_eh}
E_h = \sum_{i=1}^N \Arrowvert w_{i}V_i-P_i \Arrowvert^2
\end{equation}
in which $N$ stands for the number of corresponding pairs on silhouette, each of the $\hat{P_i}$ can be represented as the bilinear interpolation of mesh grids which contains $\hat{P_i}$, $\hat{P_i}=w_i\hat{V_i}$, in which $w_i$ remains unchanged after warping. We define the correspondence pair $\hat{P}$ and $P$ by first finding the silhouette points on the query image, then we use the 3D model as a bridge to find the corresponding 2D points on the template image. Figure ~\ref{fig:Boundary} shows the effect of this term. Note that the artifacts in red rectangle caused by misalignment of silhouette is resolved by adding the term $E_h$, which forces the face boundary of the warped template to align with the query image.

In system calibration, we obtained the 3D transformation between eyes cameras and the HMD device. After the initial alignment, we aligned the head model to the HMD to obtain the projection matrix from the head model to eye images. Therefore, we can also easily warp the NIR eye images to the query image. The eye images contain correct eye gazes and the dynamic wrinkles around eye regions that are necessary for the face synthesis. As these images have no color information, we propose a novel eye synthesis algorithm that is described in details in Section ~\ref{sec:eye_synthesis}.

\begin{figure}[ht]
  \centering
  \includegraphics[width=0.45\textwidth]{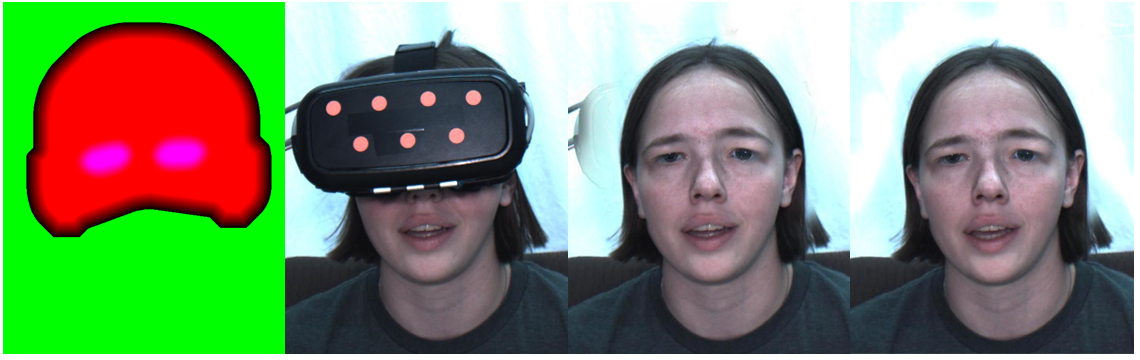}
  \caption{Illustration of blending of images. (\textit{from left to right}) (1) The mask image to blend multiple image sources. The green region represents the background we would like to keep in the final result. The red region represent the head region that is extracted from the warped reference image. The purple region is the region corresponding to the NIR eye images. Note that there is a transformation region between the red and green region. This region feather the boundary so that different image sources could be transformed smoothly from one to another. (2) The query image with the face occluded by the HMD. (3) The blending result of the query image, warped reference image, and colorized eye images. (4) The final blending result after background replacement.}
  \label{fig:blendingMask}
\end{figure}

In the final step, we blend the query image which is partially blocked by HMD, warped reference image, and two colorized eye images together. As the lower face in
the target image is often darker than faces captured under the same
illumination in the data set due to the shade of the HMD, we
first conduct a histogram transformation to adjust the reference and
two eye images to match the color of the query face image. We then blend them by using the Laplacian blending approach ~\cite{adelson1984pyramid}. Figure~\ref{fig:blendingMask} shows the formation of the final result. Note that in the rightmost image of Figure ~\ref{fig:blendingMask}, we further apply the background replacement to remove the HMD region and wires that are far from the head and not covered by the mask image.

\section{Eye Synthesis}\label{sec:eye_synthesis}
In this section, we process the warped NIR eye images in two steps. Firstly, we colorize the eye images based on the color information from the reference image. Secondly, we further refine the eye regions by removing obvious artifacts (e.g., ``red eye") during the colorization.
\subsection{Colorization}\label{sec:colorization}
We use the Lab color space as it is close to human visual conception and separates the illuminance channel from color channels. We donate the input NIR image as $I$, the reference image as $M$, and output color image as $C$. $M$ is decomposed to three channels $M_L$, $M_a$, and $M_b$. $I$ is assumed as the grayscale image for $C$. The colorization consists of two steps. We first transfer $I$ to $C_L$ based on the $M_L$. Then we transfer $M_a$ and $M_b$ to $C_a$ and $C_b$.

Two existing algorithms~\cite{reinhard2001color,shih2014style} are applied and evaluated to transfer from $I$ to $C_L$. The first algorithm~\cite{reinhard2001color} is a straightforward histogram transfer based on the standard deviations and mean values of $I$ and $M_L$. In the second algorithm~\cite{shih2014style}, the images $I$ and $M$ are aligned based on the landmarks and the SIFT flow~\cite{liu2011sift}. Then two images are decomposed into multiscale Laplacian stacks. These stacks are updated by the gain maps and are aggregated to generate $C_L$. In our problem, the performance of the second algorithm could slightly better than the first algorithm. However, it is more time-consuming due to the alignment based on the SIFT flows.

In the second step, we estimate $C_a$ and $C_b$ using the algorithm in~\cite{levin2004colorization}. The color in the channel $a$ is computed by minimizing the following energy function
\begin{equation}
E(a)=\sum_{p}((a(p)-\sum_{q\epsilon N(p)} w_{pq} a(q))^2+\alpha \sum(a(p_m)-P_m)^2
\end{equation}
where $a(p)$ is pixel $p$ on channel a, $N(p)$ is the neighbor pixel of $p$. $p_m$ and $P_m$ is the pre-defined seed pixels (\textit{i.e.}, `micro scribble' defined in~\cite{levin2004colorization}). This equation minimizes the difference between the color at pixel $p$ and its weighted averages of the neighboring pixels. The weight $w_{pq}$ is computed based on $C_L$ and statistics of the local patch around $p$. The color in the channel $b$ is also computed in the same way.

However, we need to be careful to select seed pixels. If we uniformly sample from image $M$, colors of some seed pixels could be wrong on the image $I$, such as moles and highlights. These colors propagate to following image frames gradually and generate obvious artifacts. To avoid this, we use a voting scheme to remove the unreliable seed pixels. We first run adaptive $k$-means clustering to segment the image $M$ at gray scale level. Then we only select seed pixels with high confidence that is measured by
\begin{equation}\label{eq:error}
    error = \frac{\arrowvert  I_p-I_c \arrowvert }{I_c}
\end{equation}
where $I_p$ is the intensity value of $p$th pixel in $I$ and $I_c$ is the intensity value of the center of each segment. We only select the seed colors with $error<0.06$. 

\subsection{Refinement of Eye Regions}
\begin{figure}[h]
\centering
  \includegraphics[width=3.3in]{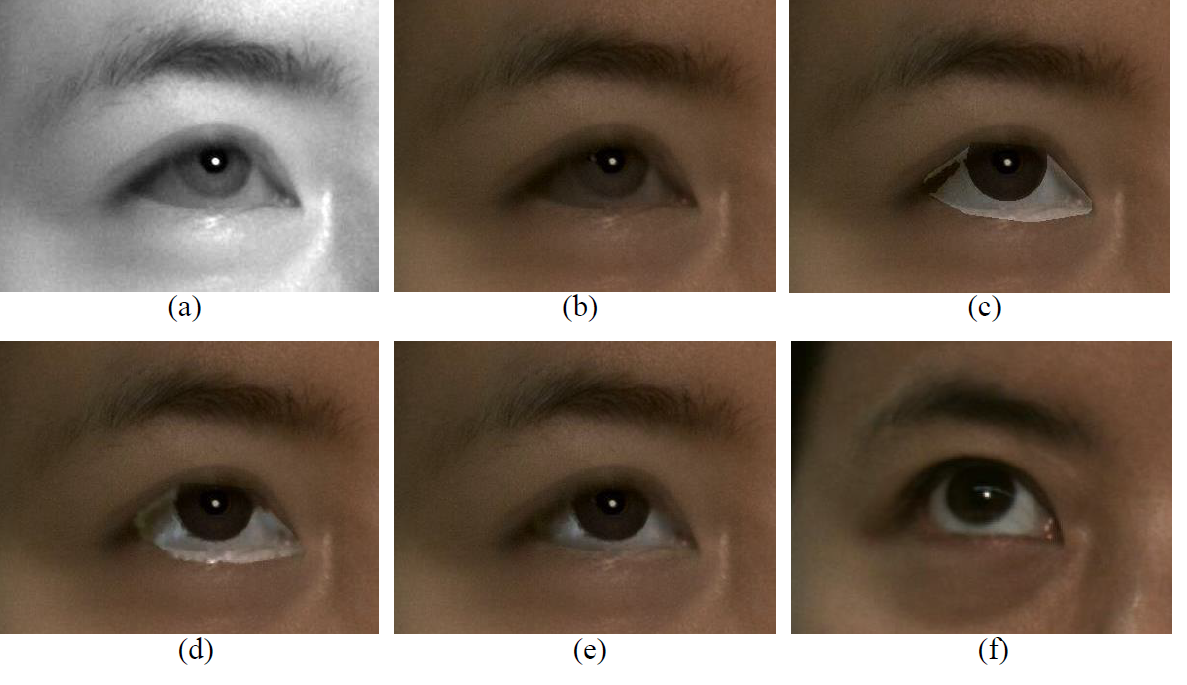}\\
  \caption{Results of eye refinement. (a) near-infrared image. (b) result after colorization (with ``red eye" effects) (c) result only using the date term $E_d$. (d) result using data and smoothness terms ($E_d$ and $E_s$). (e) result using all three terms ($E_d$, $E_s$, and $E_b$).(f) reference eye color images.  }\label{fig:eye}
\end{figure}

The eye region after colorization often contains very strong artifacts (\textit{i.e.}, ``red eye" effects) as shown in Figure~\ref{fig:eye}(b). One possible reason is that we treat the NIR image as the grayscale image. We found that, the contrast in a NIR eye image is often weaker, especially the contrast between sclera and skin and the contrast between iris and sclera. As a result, skin colors could be transferred to the regions like sclera and iris, which easily generates ``red eye" effects.

In this section, we propose an algorithm to refine the eye regions. We first detect iris and pupil boundaries in both near-infrared and color images using the intergrodifferential operator in~\cite{daugman2004iris}. Combining with the eye landmarks, we segment the eye regions into three categories, pupil, iris, and sclera. We then apply histogram transformation separately in the regions of these categories. We denote the image after histogram transformation as $C'$. This result partially removes ``red eye" effects. However, it introduces strong artifacts around boundaries of these categories and makes the result unnatural. In order to remove the artifacts, we formulate a minimization based on a cost function with three terms.

\begin{equation}
E_d= \sum_{p_m}(C''_L(p_m)-C'_L(p_m))^2
\end{equation}

\begin{equation}
E_s=\sum_{p}((C''_L(p)-\sum_{q\epsilon N(p)} w_{pq} C''_L(q))^2
\end{equation}

\begin{equation}
E_b= \sum\limits_{p} (\arrowvert N_p \arrowvert C''_L(p)-\sum\limits_{q\in \Omega}C''_L(q)-\sum\limits_{q\not \in \Omega}C_L(q)-\sum\limits_{q \epsilon N_p} V_{pq})^2
\end{equation}

\begin{equation}\label{eq:cost2}
E=E_d+\alpha_1 E_s+ \alpha_2 E_b
\end{equation}

where $C''_L$ is the L channel of the output eye image $C''$ (the same procedure is also applied to a and b channels), $p_m$ is a seed pixel (the seed colors are selected using the criteria described in Equation~\ref{eq:error}), $C'_L(p_m)$ and $C''_L(p_m)$ are values of the L channel on pixel $p_m$ for input image $C'_L$ and output image $C''_L$ respectively, $N_p$ is neighboring pixels of pixel $p$, $\arrowvert N_p \arrowvert$ is the number of $N_p$, $\Omega$ is the mask that includes only the eye region, and $V_{pq}=C_L(p)-C_L(q)$ is the gradient value of this two pixels. $\alpha_1$ and $\alpha_2$ are tuned based on our experiments. The weight $w_{pq}$ is proportional to the normalized correlation between two values of the L channels. $w_{pq}$ is given by
\begin{equation}
w_{pq}=1+\frac{1}{\sigma_p^2}(C_L(p)-\mu_p)(C_L(q)-\mu_p)
\end{equation}
where $\mu_p$ and $\sigma_p$ are the mean and standard deviation of pixel values in an image patch around $p$.

The first term $E_d$ is the data term that color an unknown pixel same as the seed pixel in the input image. $E_s$ is the smoothness term that makes the color are smoothly transformed among its neighborhood. The last term $E_b$ is the boundary term that is inspired by the gradient image editing~\cite{perez2003poisson}. This term is equivalent to the Poisson equation with Dirichlet boundary conditions. The refinement algorithm is summarized below.

\begin{algorithm}
\DontPrintSemicolon
\KwData{The eye image $C$ that is the color image after colorization and contains ``red eye effects".}
\KwResult{Refined eye image $C''$.}

\begin{enumerate}
  \item Detection of iris and pupil boundaries based on the intergrodifferential operator in~\cite{daugman2004iris}.
  \item Segmentation of pupil, iris, and sclera using eye landmarks and boundaries of iris and pupil.
  \item Histogram transformation for each category and each color channel. $C'$ is denoted as the output after transformation.
  \item Selection of seed pixels $p_m$ based on Equation~\ref{eq:error}.
  \item Minimization of Equation~\ref{eq:cost2}.
\end{enumerate}

\caption{Refinement of eye regions.\label{alg:pose}}
\end{algorithm}

Figure~\ref{fig:eye} shows one group of eye images after refinement. In this Figure, we can find that image (b) contains ``red eye" effects, image (c) has strong artifacts around boundaries of segmentation, and image (e) is the result using all three terms and ``red eye" effects are removed.

\section{Experiment}\label{sec:exp}
Our system framework is tested on both simulation and mobile setups. For the simulation setup, the goal is to validate and quantify the accuracy of our system.




 \subsection{Evaluation of 3D reconstruction}
\begin{figure}
\centering
  \includegraphics[width=0.45\textwidth]{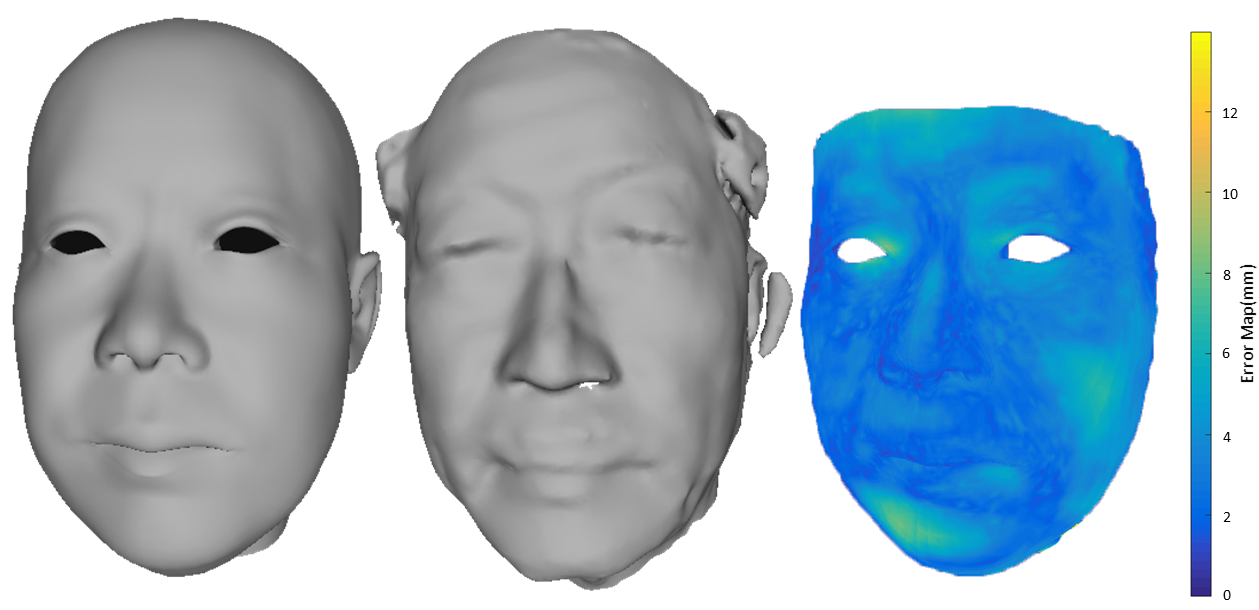}\\
  \caption{Evaluation of 3D reconstruction. \textit{from left to right} (1) The head model reconstructed by our algorithm. (2) The model scanned by a structured light 3D scanning system. This model serves as ground truth. (3) The error map between our model and the ground truth.}\label{fig:error mapReconstruction}
\end{figure}

 We first scan 3D head models with a high resolution structured light 3D scanning system which has everage reconstructing error less than 2mm, more details about this scanner could be found in ~\cite{fullbody2015full}. These 3D models serve as ground truth in our evaluation. To measure the difference between our models and ground truth, we first align them by using facial 3D landmarks and then the ICP algorithm ~\cite{besl1992method} is used to refine the alignments. In order to find the dense correspondences between our model and the ground truth, we search the nearest point on the ground truth mesh surface along the normal direction of each 3D point on the model we reconstructed. The Euclidean distances are then computed for all the pairs of 3D points. The mean error distance between our model and the ground truth is 2.926 mm. Figure~\ref{fig:error mapReconstruction} shows one example of our model, the corresponding ground truth, and the error map.
 

 \subsection{Evaluation of Face Tracking}
\begin{figure}
\centering
  \includegraphics[width=0.46\textwidth]{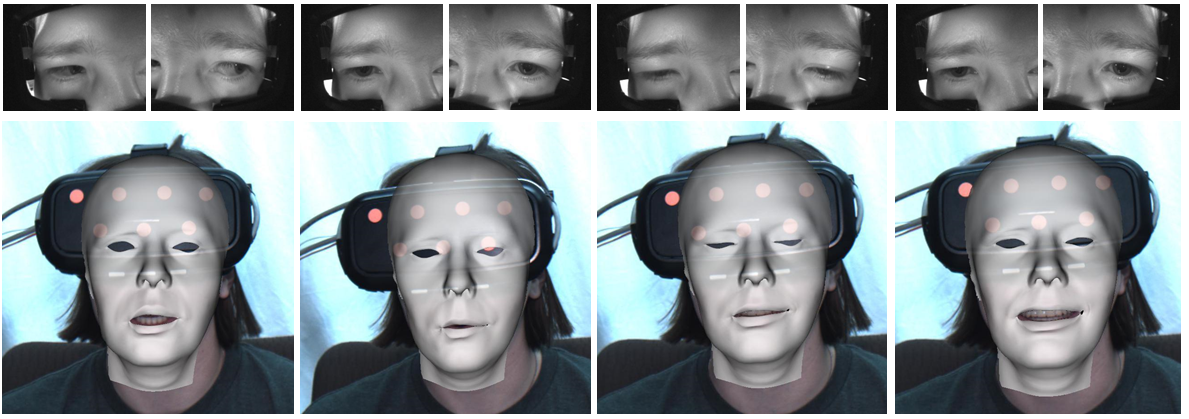}\\
  \caption{Evaluation of tracking of 3D face models. The 3D model is overlaid on the original input frame. The various facial expression, especially the eye blinks, are tracked robustly by our algorithm.}\label{fig: trackingresult}
\end{figure}

\begin{figure}
\centering
  \includegraphics[width=0.47\textwidth]{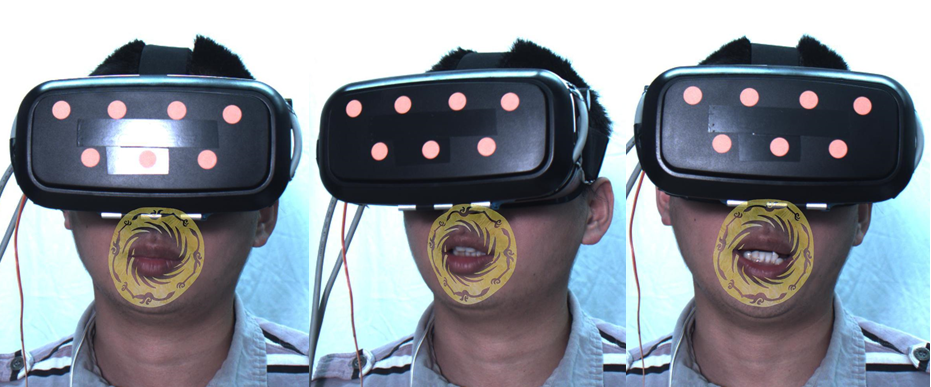}\\
  \caption{Evaluation of tracking performance with a projected virtual pattern. The virtual pattern is deformed smoothly and consistently with the mouth movements.}\label{fig:TrackingResultPose}
\end{figure}

Our algorithm described in section~\ref{sec:alignment} can robustly track 3D face models in real-time. As shown in Figure~\ref{fig: trackingresult}, these models could contain various facial expressions, such as eye blinks and mouth movements. In order to further evaluate the tracking performance, we project a virtual pattern to each input image frame. As shown in Figure~\ref{fig:TrackingResultPose}, we can find that the virtual pattern is deformed smoothly and consistently with various mouth movements.\textbf{ Video of Tracking results can be found in supplementary material.}
 
 \subsection{Evaluation of Eye Synthesis}
\begin{figure*}
\centering
  \includegraphics[width=0.85\textwidth]{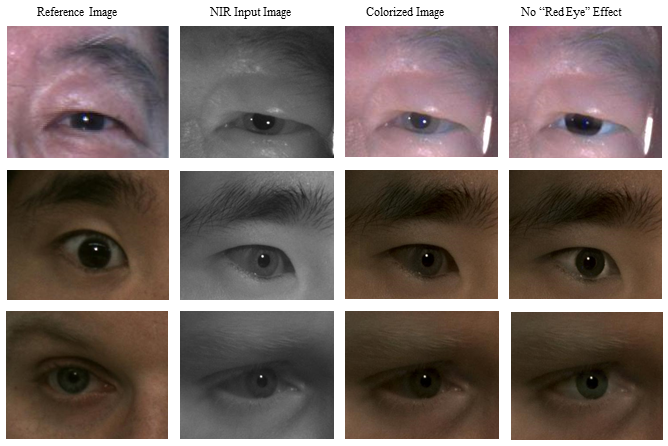}\\
  \caption{Results of eye synthesis. (\textit{1st column}) Reference images retrieved from pre-recorded image frames. (\textit{2st column}) Input of NIR images.(\textit{2nd column}) Results after colorization. (\textit{3rd column}) Results after refinement of eye regions.}\label{fig:eyeresults}
\end{figure*}

It is essential to generate accurate eye movements in the final synthesized face image. Figure~\ref{fig:eyeresults} demonstrates our results of eye synthesis for three different users. The first column of images are the reference images retrieved from the pre-recorded image frames. The second column of images contains the input NIR images. The third column contains images after colorization that is the first step of our eye synthesis. We can find that the color appearances have been adjusted to be very similar to the reference images. However, the ``red eye" effects are also quite obvious in these images. After the second step of our eye synthesis algorithm, the ``red eye" effects are removed and more realistic eye images are generated (the fourth column). 
 

\subsection{Runtime}

Overall, our implementation on CPU takes around 560 ms to process one frame on  Intel Core i7-4710 CPU(3.4GHz) with the color image in resolution of $1280 \times 960 $ and two NIR images in resolution of $640 \times 480$. Table ~\ref{tab:table1} shows the runtime of major components in our system. The face synthesis component consists of reference image retrieval, face warping and final blending, which is the most time consuming modules. We believe that, by using the parallel processing power of GPUs and reducing the image resolution, we can achieve real-time performance.

\begin{table}[h!]
  \centering
  \caption{Runtime of Different Modules}
  \label{tab:table1}
  \begin{tabular*}{0.35\textwidth}{ccc}

    \toprule
    Tracking
 & Eye Colorization
&Face Synthesis
\\
    \midrule
    8ms & 160ms &400ms\\

    \bottomrule
  \end{tabular*}
\end{table}

\subsection{Face Synthesis Results}

\begin{figure*}
\centering
  \includegraphics[width=0.85\textwidth]{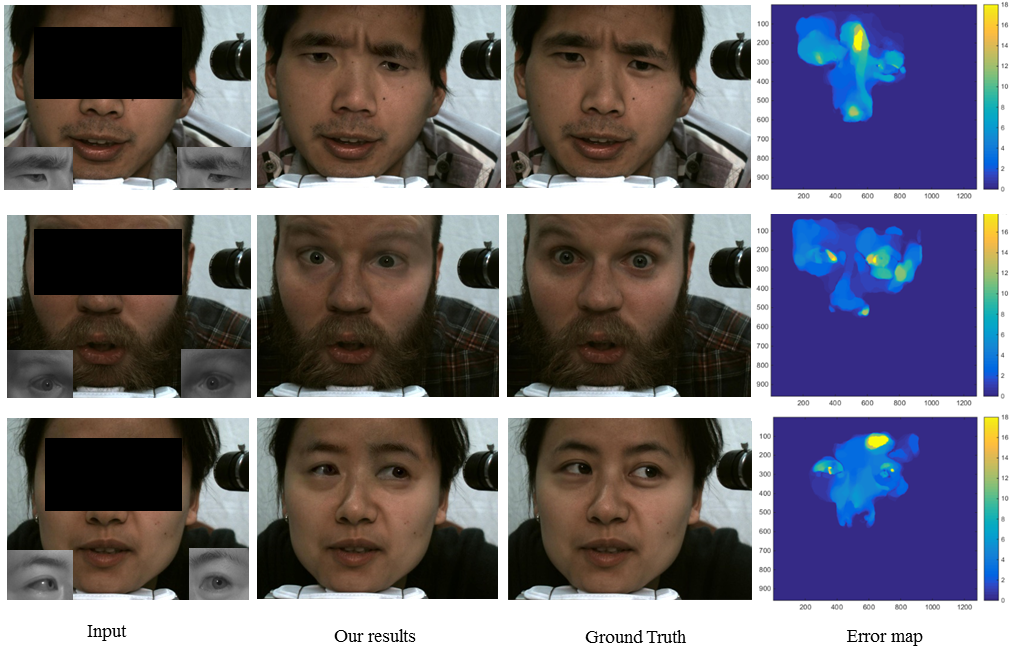}\\
  \caption{Simulation Results. (\textit{1st column}) have the input images with eye images shown at the bottom of each face image. (\textit{2st column}) contains synthesized face images by our system. (\textit{3st column}) have the ground truth images.(\textit{4st column}) is the error map between ground truth and synthesized image.}\label{fig:simulation}
\end{figure*}

As the ground truth is available in the simulation setup, we can evaluate our expression tracking and eye colorization algorithm by computing the error map between our synthesized image and the ground truth. Figure~\ref{fig:simulation} shows results for different users. The average intensity error is around 5.6 under the area of mask based on intensity range from ${0 \sim 255}$, which indicates that our eye colorization algorithm can produce accurate results. In the simulation setup, the head pose is fixed. We ask users to perform as much expression as they can in an off-line expression database. We calculate the expression weights by using facial landmarks for all the frames in the database. For the query frame with upper face occluded, we also calculate the expression parameters by using our algorithm, then retrieving for the best matched expression in database. This retrieved image is used to fill the missing part of the query image. Note that in Figure~\ref{fig:simulation}, the facial details are reconstructed by using the best matched expression template in database, this demonstrates the effectiveness of our expression tracking algorithm.

\begin{figure}[ht]
\centering
  \includegraphics[width=0.45\textwidth]{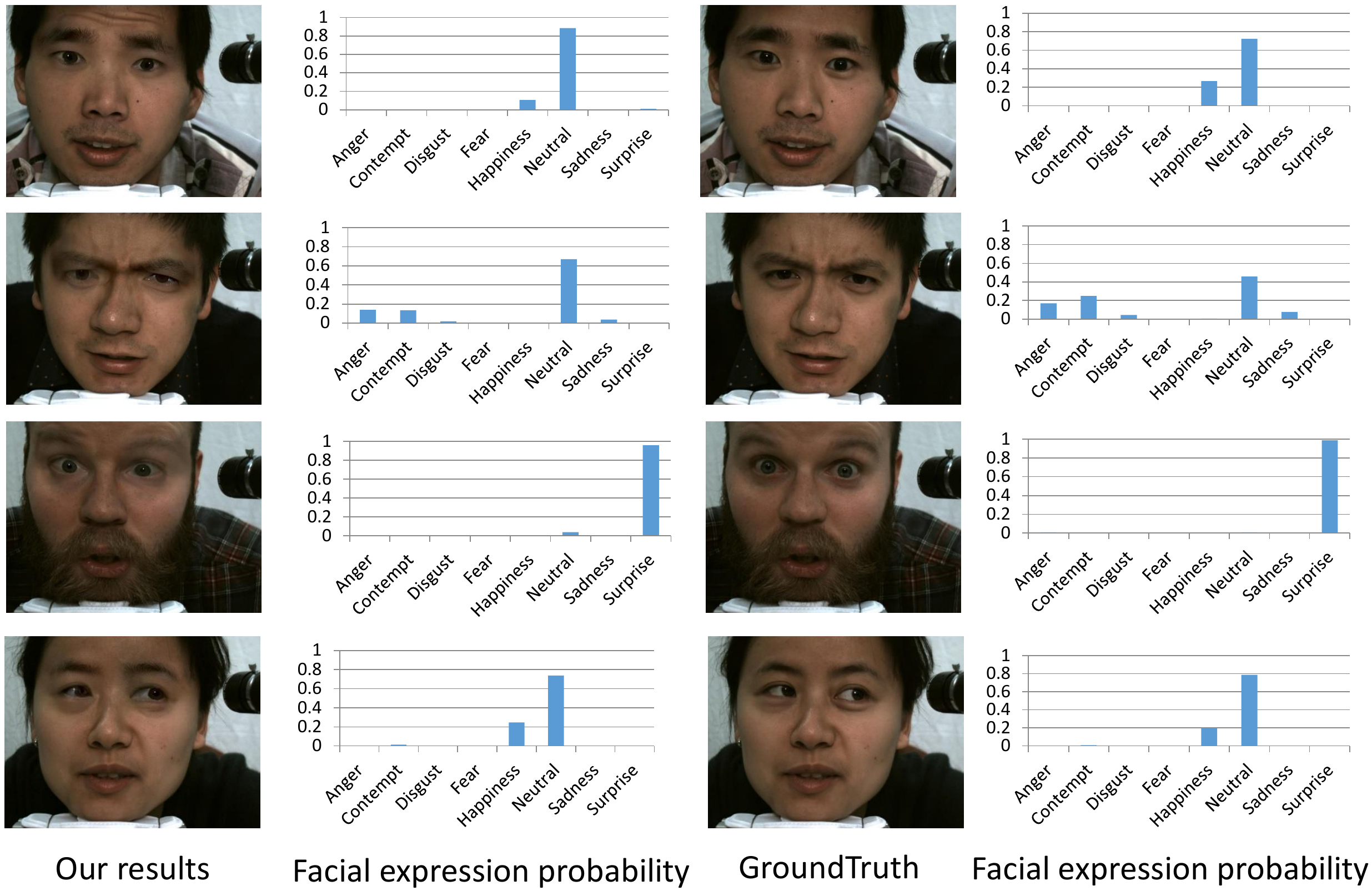}\\
  \caption{Comparison on expression recognition between our results and ground truth. }\label{fig:Facialexpression}
\end{figure}

To further validate our system with some quantitative measurements, we apply facial expression recognition on Microsoft Emotion API ~\cite{api} on both our results and ground truth images. The results are shown in Figure~\ref{fig:Facialexpression}. Expressions are divided in 8 different categories (i.e., anger, contempt, disgust, fear, happiness, neutral, sadness, and surprise). This experiment shows that our system can generate accurate facial expressions.

\begin{figure*}
\centering
  \includegraphics[width=0.69\textwidth]{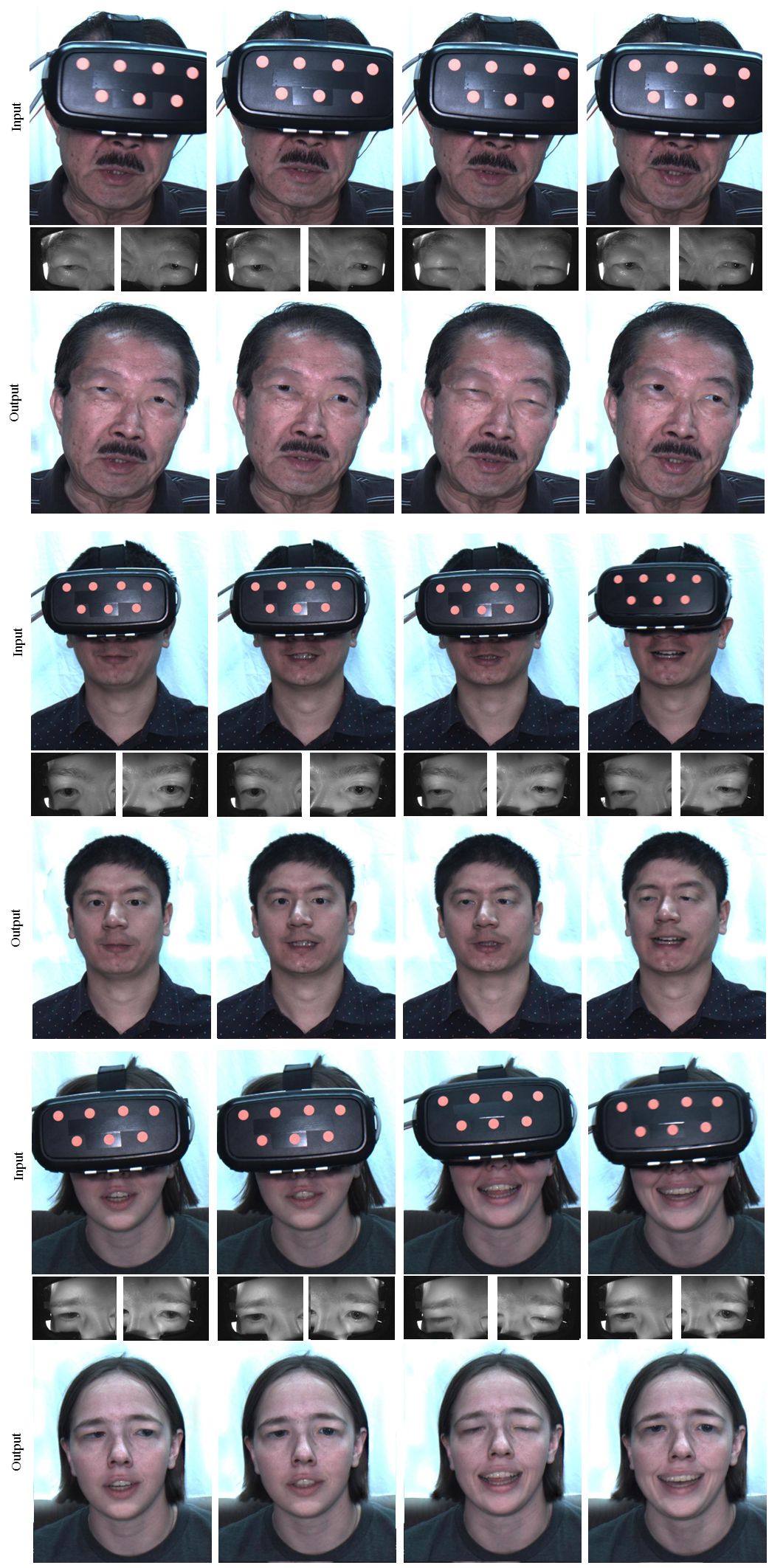}\\
  \caption{Results of our system. }\label{fig:FinalResults}
\end{figure*}

Figure~\ref{fig:FinalResults} shows results for our mobile setup. We have tested our system on different users with various facial expressions including eye and month movements. Our results demonstrate the effectiveness and robustness of our system. \textbf{More results including videos can be found in our supplementary material}.


\section{Conclusions}\label{sec:conc}
In this paper, we tackle the face synthesis problem in which the upper face region is severely occluded by the HMD. We design a novel system framework that consists of two NIR cameras capturing the eye regions and one visible-light camera to capture the face image with only lower part visible. In order to synthesize realistic face images, we present two novel algorithms in our framework. Firstly, we propose a novel algorithm to align and track 3D head model based on the input image with a large portion of face occluded by the HMD. Secondly, a novel approach to synthesize eye regions is presented. In this approach, we colorize the NIR eye images and further remove the ``red eye" effects.

In the future, we would like to explore other directions to further improve our framework. For example, we could apply different alternates to speed up the framework and achieve real-time performance with modern GPU.



\section*{Acknowledgements}

Thank Chen Cao and Kun Zhou to share their face datasets. The research is funded in part by US National Science Foundation grants IIS-1231545 and IIP-1543172, Natural Science Foundation of China (No.61332017, 61572243) and Zhejiang Provincial Natural Science Foundation grant LY13F020019.

\bibliographystyle{acmsiggraph}

\nocite{*}
\bibliography{template}

\end{document}